\documentclass[review]{elsarticle}

\usepackage{hyperref}

\usepackage{booktabs}
\usepackage{multirow}
\usepackage{algorithm}
\usepackage{algpseudocode}

\usepackage{graphicx}

\journal{Journal of \LaTeX\ Templates}

\begin{document}

\begin{frontmatter}

\title{ISEC:\\Iterative over-Segmentation via Edge Clustering}

\cortext[corr]{Corresponding author}
\author{Marcelo Mendon\c{c}a}
\author{Luciano Oliveira}
\ead{lrebouca@ufba.br}
\address{Intelligent Vision Research Lab, \\ Federal University of Bahia, \\ Salvador-BA, Brazil}

\begin{abstract}
Several image pattern recognition tasks rely on superpixel generation as a fundamental step. Image analysis based on superpixels facilitates domain-specific applications, also speeding up the overall processing time of the task. Recent superpixel methods have been designed to fit boundary adherence, usually regulating the size and shape of each superpixel in order to mitigate the occurrence of undersegmentation failures. Superpixel regularity and compactness sometimes imposes an excessive number of segments in the image, which ultimately decreases the efficiency of the final segmentation, specially in video segmentation. We propose here a novel method to generate superpixels, called iterative over-segmentation via edge clustering (ISEC), which addresses the over-segmentation problem from a different perspective in contrast to recent state-of-the-art approaches. ISEC iteratively clusters edges extracted from the image objects, providing adaptive superpixels in size, shape and quantity, while preserving suitable adherence to the real object boundaries. All this is achieved at a very low computational cost. Experiments show that ISEC stands out from existing methods, meeting a favorable balance between segmentation stability and accurate representation of motion discontinuities, which are features specially suitable to video segmentation.
\end{abstract}


\end{frontmatter}


\section{Introduction}

The result of an image over-segmentation lies in somewhere between the pixel and the object segmentation itself. The term ``superpixel'' has been used to refer to the information in such intermediate level. Indeed, many tasks that involves image segmentation take advantage of working with superpixels rather than pixels. Examples range from tracking \cite{SpxTrack} through 3D reconstruction \cite{Spx3DReconst} to semantic image segmentation \cite{SpxCNN}. Superpixels are also applied in many domain-specific applications, such as in traffic analysis \cite{SpxTraffic}, and in biological \cite{SpxBiological} and medical \cite{SpxMedical} image segmentation, just to name a few. In view of this wide applicability, many superpixel methods have been proposed in recent years (see recent surveys in \cite{SpxSurveyStutz,SpxSurveyWang}). 

Research on superpixel started with image segmentation methods running on over-segmentation mode. Earlier methods were not explicitly conceived to generate image superpixels. For instance, segmentation methods such as watershed (WS) \cite{WS} and normalized cuts (NC) \cite{NC} are able to produce the over-segmentation effect by adjusting the number of seeds, in the case of the former, or the number of graph partitions, in the latter. Later methods such as edge augmented mean shift (EAMS) \cite{EAMS}, Felzenszwalb-Huttenlocher (FH) \cite{FH} and quick shift (QS) \cite{QS} were specifically designed to produce image over-segmentation by pursuing adherence to real object contours. In these latter methods, the rationale consists in automatically regulating the amount of generated superpixels according to internal parameters, although without providing direct control over the number of generated segments. More recently, several methods have been groundly developed towards generating superpixels by directly providing control over the number of segments \cite{VEK, SLIC, ERS, SEEDS, CWS, LSC}; the goal is to divide the image into a generally user-defined fixed number of segments, also trying to match the segment boundaries as accurate as possible to the real object contours. The advantages of these methods are (i) the larger the chosen number of superpixels, the lower the probability of occurring undersegmentation errors, \textit{i.e.}, when the same segment overlaps different objects, (ii) approaches based on a fixed number of segments usually provide more regular superpixels regarding shape and size, which can be useful for some applications, \textit{e.g.}, object recognition \cite{SpxObjRecog} and labeling \cite{SpxLabeling}, and (iii), since superpixel methods are commonly used as a preprocessing stage for other computer vision tasks, by keeping control over the number of segments, one can make the computational effort in the further steps be more predictable.

\begin{figure}[!t]
\centering
\includegraphics[width=1\textwidth]{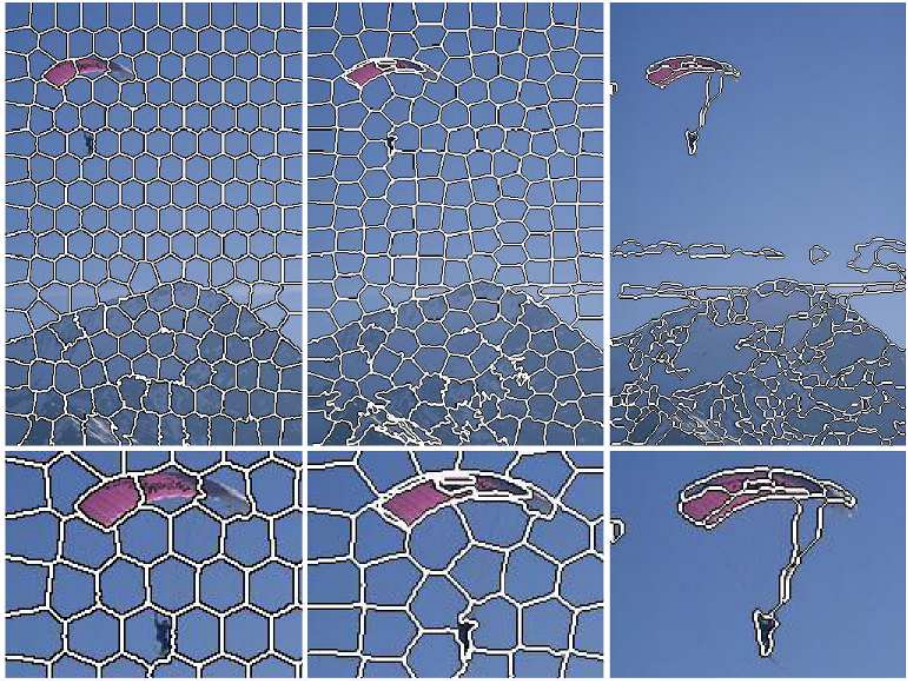}
\caption{Examples of superpixel segmentation. From left to right column: SLIC \cite{SLIC} segmentation, with $k=200$, LSC \cite{LSC} segmentation, with $k=200$, and ISEC segmentation, with $k=158$, where $k$ is the number of generated superpixels; for SLIC and LSC, $k$ is user-defined, while for ISEC, $k$ is image-based adaptive. Bottom images highlight the parachute man segmentation for each method: SLIC and LSC present undersegmentation failures, while ISEC correctly segments the image object.}
\label{fig_parachutes}
\end{figure}

Deciding about the better strategy to generate superpixels -- if the number of segments is automatically or externally provided -- gives rise to a straightforward question: ``How many superpixels should an over-segmentation method generate for a given image?'' We could intuitively say that the answer depends on the image. In words, if there are many objects, more segments will be necessary; otherwise, just a small number would be enough. Despite this immediate conclusion, more recent superpixel methods have not actually addressed the over-segmentation problem from such an adaptive perspective. Specially because of the undersegmentation problem, most of the recent methods goes towards generating a fixed and externally provided number of superpixels. To illustrate that situation, Fig. \ref{fig_parachutes} shows the results of superpixel generation by two state-of-the-art methods: simple linear iterative clustering (SLIC) \cite{SLIC} (left column), and linear spectral clustering (LSC) \cite{LSC} (middle column). In the figure, both SLIC and LSC were set to yield 200 superpixels. With this number, both methods provide superpixels that correctly adhere to the contours of the big mountain in the figure, although the parachute man is not accurately segmented, causing an undersegmentation failure. Errors, like in the parachute man segmentation, mainly take place when the generated superpixels are much bigger than the object being segmented. An alternative to tackle this problem could be to increase the number of superpixels, making them smaller. However, an excessive over-segmentation implies to segment other objects in the image into much more parts than necessary, reducing efficiency\footnote{In this context, efficiency means capturing the maximum number of real object contours with the minimum number of superpixels.}. Also, the grid-like segmentation provided by methods such as SLIC and LSC does not allow to distinguish homogeneous areas from object-filled regions in the image (\textit{e.g.}, the blue sky in the background), resulting in unnecessary over-segmentation of these areas.

\subsection{Contributions}

A method called iterative over-segmentation via edge clustering (ISEC) is proposed in this paper. ISEC is a novel superpixel method that addresses the over-segmentation problem from a different perspective in contrast to recent state-of-the-art approaches. The proposed method iteratively clusters edges extracted from the image objects, providing superpixels that are, at the same time, adaptive in size, shape and quantity, while preserving efficient adherence to the real object boundaries. All this is achieved at a very low computational cost. The right column of Fig. \ref{fig_parachutes} illustrates the result of ISEC over-segmentation on the same image segmented by SLIC and LSC. It is noteworthy that ISEC is able to self-adapting to correctly segment both the big mountain and the small parachute man, avoiding unnecessarily segmenting most of the background comprised of the homogeneous blue sky. Thanks to the adaptiveness of the proposed method, the resulting over-segmentation in Fig. \ref{fig_parachutes} was accomplished with comparatively fewer (158) superpixels than SLIC and LSC, turning ISEC to be more efficient in this case. On the other hand, superpixels generated by SLIC and LSC are much more regular in shape and size, while ISEC prioritizes superpixels more adjusted to the real object contours. For some applications, such as that one to segment magnetic resonance images \cite{SpxMedical_2}, adherence to real object contour is a crucial characteristic rather than superpixel regularity or compactness, since superpixels with compact structure are less able to cover complete objects in that type of image.

The proposed method iteratively groups image edges in order to form clusters. The rationale is that the borders that delimit these clusters are to be superpixels presenting real object contour adherence efficiently (see Section \ref{sec_propMethod} for details). Results from video-based evaluation metrics \cite{SPXBenchmark} indicate that the superpixels generated by ISEC demonstrate promising performance regarding video segmentation. This assessment is corroborated by the results found in a video object segmentation experiment (see Section \ref{sec_experiments}). Indeed, previous comparisons \cite{ComparisonSpxVideoSeg} have demonstrated that superpixels with characteristics similar to ISEC perform better than grid-like, regular-shaped superpixels in video object segmentation. In \cite{SPXBenchmark}, Neubert and Protzel make a thorough evaluation on superpixel methods focused on video segmentation; they point out a lack of superpixel methods able to produce segmentations that are consistent in two criteria: segmentation stability and accurate representation of motion discontinuities. Our experiments show that ISEC stands out from existing methods, meeting a favorable balance between the aforementioned criteria. 



\section{Related works} \label{sec_relWork}

Some of the existing superpixel methods are reviewed here. The presented methods are further compared to ISEC in our experimental evaluation (results on evaluation performance are reported in Section \ref{sec_experiments}).

\bigskip\noindent \textbf{WS} (1992) \cite{WS} -- \textit{\underline{W}ater\underline{s}hed} is an well-known, very fast over-segmentation algorithm. It works by iteratively growing user-defined seeds until they reach the borders of other adjacent growing seeds. The number of superpixels is determined by the number of seeds. There is an open source C/C++ implementation available at OpenCV\footnote{\url{http://opencv.willowgarage.com/wiki/}}.

\smallskip\noindent \textbf{NC} (2000) \cite{NC} -- The \textit{\underline{N}ormalized \underline{C}uts} is a widely used graph-based algorithm, and maybe the first applied to generate superpixels \cite{SpxSurveyStutz}. As such, NC recursively produces a graph, in which the vertices correspond to the image pixels, while the weights account for similarities among the pixels. Superpixels are obtained by partitioning this graph in a user-defined number of segments. This process takes a lot of time when compared to other methods. Our experiments using NC have been done with resized images (160 pixels on the longer side, while keeping the aspect ratio) to save time. An open source implementation for Matlab and C++ is available\footnote{\url{http://www.timotheecour.com/software/ncut/ncut.html}}.

\smallskip\noindent \textbf{EAMS} (2001) \cite{EAMS} -- \textit{\underline{E}dge \underline{A}ugmented \underline{M}ean \underline{S}hift} is a density-based algorithm that performs an iterative mode-seeking procedure in a computed density image. This procedure consists in finding modes in color or intensity feature space so that each pixel is assigned to the corresponding mode where it falls into. The superpixels are composed by the pixels that converge to the same mode. This method mainly differs from the k-means algorithm since it does not require a user-defined number of clusters. Source codes in C and Matlab are available\footnote{\url{http://www.wisdom.weizmann.ac.il/~bagon/matlab.html}}.

\smallskip\noindent \textbf{FH} (2004) \cite{FH} -- \textit{\underline{F}elzenszwalb-\underline{H}uttenlocher} is a graph-based superpixel approach. The algorithm measures the evidence of a boundary between two regions by computing a graph-based representation of the image. FH tries to preserve details when a region presents low variability, while details in image regions with high variability are ignored. This is done by using a greedy algorithm that also satisfies global properties. FH automatically controls the number of superpixels, and it is among the fastest superpixel methods. An open source C++ implementation is available\footnote{\url{http://www.cs.brown.edu/~pff/segment/}}.

\smallskip\noindent \textbf{QS} (2008) \cite{QS} -- \textit{\underline{Q}uick \underline{S}hift} is also a mode-seeking segmentation method. Instead of relying on a gradient ascent approach such as EAMS, QS uses a medoid shift. The medoid is faster than the mean shift in Euclidean space. The algorithm builds a tree of paths by moving each point in the feature space to the nearest neighbor that increases the density estimative. Superpixels are obtained by splitting the branches of the tree using a threshold. The number of segments is automatically controlled, and the method is not fast. Implementation in C and Matlab is available as part of the VLFeat library\footnote{\url{http://www.vlfeat.org/overview/quickshift.html}}.

\smallskip\noindent \textbf{VEK} (2010) \cite{VEK} -- \textit{\underline{Vek}sler Superpixels}. This method is also a graph-based approach, but based on regular partition. The segmentation is assumed as an energy minimization problem that explicitly encourages constant intensity inside the superpixels. Segmentations are generated by gluing overlapping image patches until each pixel remains only inside of an individual patch. There are two available implementations of VEK, one for compact and other for constant-intensity superpixels. In our experiments, we evaluated only the latter\footnote{\url{http://www.csd.uwo.ca/faculty/olga/code.html}}. VEK is among the slowest algorithms, and the number of superpixels is user-defined.

\smallskip\noindent \textbf{SLIC} (2010) \cite{SLIC} -- \textit{\underline{S}imple \underline{L}inear \underline{I}terative \underline{C}lustering} is inspired by the k-means algorithm. However, the strategy is modified to limit the region in which the algorithm searches for similarities among the pixels. SLIC considers only the pixels within a defined spatial neighborhood. The process begins by dividing the image into a regular grid where the number of cells is defined by the user. Color and spatial distances among the pixels are measured in order to iteratively adjust the cells to form superpixels. SLIC is among the fastest algorithms. An implementation in C++ and Matlab is available\footnote{\url{http://ivrl.epfl.ch/supplementary_material/RK_SLICSuperpixels/index.html}}.

\smallskip\noindent \textbf{ERS} (2011) \cite{ERS} -- \textit{\underline{E}ntropy \underline{R}ate \underline{S}uperpixel Segmentation} is formulated as a maximization problem on a graph-based image representation. The method relies on an objective function for superpixel segmentation that accounts for the entropy among the graph edges and the size of the segments. ERS generates superpixels by selecting the edges that yield the largest gain of the objective function. ERS iterates until the user-defined number of clusters is reached, which usually takes over a second. An implementation in C/C++ is available\footnote{\url{https://github.com/mingyuliutw/EntropyRateSuperpixel}}. 

\smallskip\noindent \textbf{SEEDS} (2012) \cite{SEEDS} -- \textit{\underline{S}uperpixels \underline{E}xtracted via \underline{E}nergy-\underline{D}riven \underline{S}ampling}. This method iteratively optimizes an energy function, starting from a grid-like image segmentation and then modifying the boundaries according to the function. In each step, pixels are exchanged between neighboring segments so that if the new composition produces a gain of the energy function, it is retained, otherwise it is rejected. The number of generated superpixels is user-defined, and the method is among the fastest ones. An implementation in C and Matlab is available\footnote{\url{http://www.mvdblive.org/seeds/}}.

\smallskip\noindent \textbf{CWS} (2014) \cite{CWS} -- \textit{\underline{C}ompact \underline{W}ater\underline{s}hed} is a modified version of the original WS method, aiming at providing control over the superpixel compactness. The modification encompasses two aspects: grid arrangement of the seeds and incorporation of a controllable compactness constraint. This is done by means of a single parameter that controls the level of compactness regardless the number of segments or the image size. CWS is so fast as the original WS. An implementation in C/C++ is available\footnote{\url{https://www.tu-chemnitz.de/etit/proaut/forschung/cv/segmentation.html.en}}.

\smallskip\noindent \textbf{LSC} (2015) \cite{LSC} -- \textit{\underline{L}inear \underline{S}pectral \underline{C}lustering} uses kernel functions to construct a ten-dimensional representation of the image. This high dimensionality is given by a set of features extracted for each pixel. After initializing clusters from seeds distributed along the image, an iterative process measures the distances between pixels and clusters in the feature space, assigning the pixels to the cluster with the closest distance. This process ends when the clusters become stable, which is accomplished in a few reasonable time. The number of clusters (superpixels) is user-defined. An implementation in C++ with Matlab interface is available\footnote{\url{https://jschenthu.weebly.com/projects.html}}.



\begin{figure}[!t]
\centering
\includegraphics[width=\textwidth]{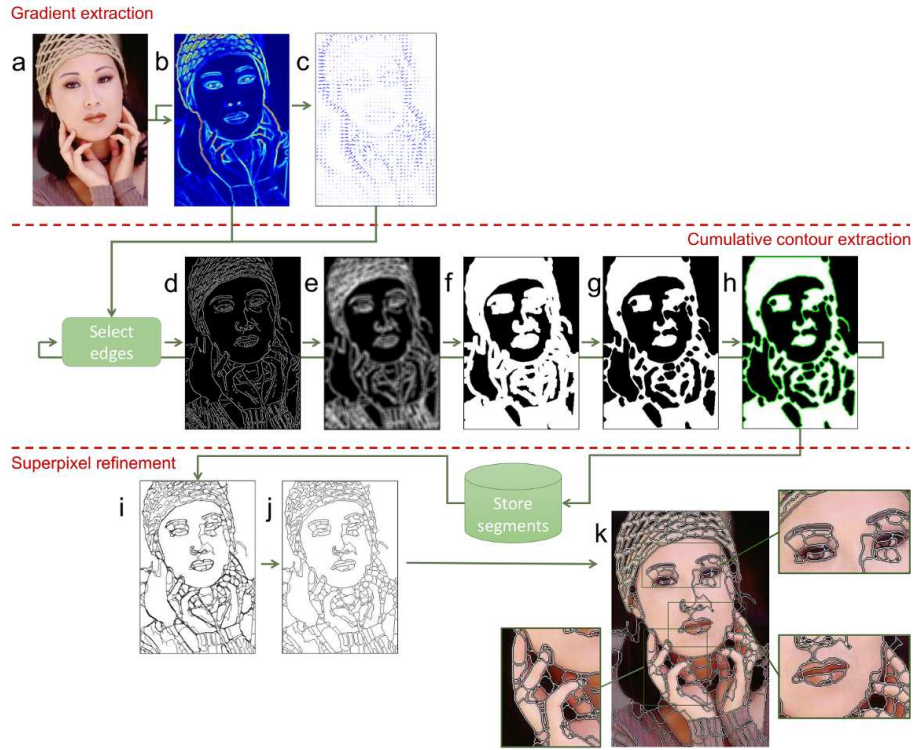}
\caption{Top-down view of ISEC. Gradient extraction: An input image (a) is used to compute gradient magnitudes (b) and orientations (c) in the $x$ and $y$ axes. Cumulative contour extraction: For each iteration, an edge set (d) is selected from the gradient; the edges are stretched (e) by edge density filtering, being binarized to form clusters (f); a thinning operation is performed on the clusters to readjust their shapes (g); the borders of the clusters are extracted (h) and stored. Superpixel refinement: The accumulated segments (i) are refined to produce the final result (j). The generated superpixels are showed over the input image (k); some parts are zoomed to highlight segmentation details.}
\label{fig_isec_outline}
\end{figure}

\section{Superpixel generation via edge clustering} \label{sec_propMethod}

Superpixels are generated from image edges, which are iteratively selected and grouped to form clusters. Clustered pixels represent image locations where the spatial neighborhood is densely filled by edges. Since the edges are strongly related to the objects contours and texture, the clusters resemble the object shapes. Superpixels are ultimately obtained by extracting the borders that separate these clusters from image regions where there are no edges. By adjusting the edge selection procedure so that a group of edges are selected at the beginning of the process, and edges are progressively removed at each new iteration, the input image is over-segmented from the outer contours of the objects to their small internal parts. Figure \ref{fig_isec_outline} depicts the top-down view of ISEC. Next, each part of the proposed method is described.

\subsection{Gradient extraction} \label{subsec_preproc}

Given an input image (Fig. \ref{fig_isec_outline}.a), the first step is to compute, for each pixel, the gradient magnitude (Fig. \ref{fig_isec_outline}.b) and orientation along the $x$ and $y$ axes (Fig. \ref{fig_isec_outline}.c). Canny edge detector is used to calculate the image gradients. The gradients are individually computed over each RGB channel. The final gradient magnitude is given by the highest magnitude presented by the pixel among the channels. By doing that, color information is also encoded in the edge detection process.

\subsection{Cumulative contour extraction} \label{subsec_iterative}

Once the image gradients are computed, a set of edges (Fig. \ref{fig_isec_outline}.d) is initially selected from the gradient map. Edge selection is iteratively performed by combining non-maximum suppression and double thresholding, just as in Canny detection\footnote{Different edge detection strategies could be used instead, and Canny was chosen because of the best results at the experiments.}. Initially, small thresholds are used so that a great amount of edges are selected. As the process iterates, the thresholds become more restrictive, yielding progressively fewer edges. The idea consists in extracting the object contours in an outside-in manner, \textit{i.e.}, starting from the silhouette to inward subparts. This strategy is accomplished by handling the edge detection thresholds. For small thresholds, many edges are selected, including those ones related to strong contours (high gradient magnitude), or even edges originated from smooth contours and textures (low gradient magnitude). This initial great amount of edges tend to compose clusters with the same shape as the object that originated the edges. As the process iterates, the thresholds are incremented, becoming increasingly restrictive. These higher thresholds prevent many edges to be selected from smooth contours and texture, remaining only the edges related to stronger contours. These stronger edges tend to resemble smaller internal subparts of bigger objects.

In principle, directly using edges to construct superpixel is tempting, since edges are strongly related to the real object contours. However, this is not practical because edges generally present discontinuities, resulting in unclosed segments. To tackle this problem we perform a spatial linear filtering on the edge image, given by

\begin{equation}
	ED = \frac{1}{S^{2}}\sum_{i=1}^{S^{2}}p_{i} \, ,
\label{eq:edge_density}
\end{equation}
where $ED$ denotes edge density, which is calculated by an arithmetic mean over the neighborhood $S\mbox{x}S$ of the pixel $p_i$.

\begin{figure}[!t]
\centering
\includegraphics[width=0.8\textwidth]{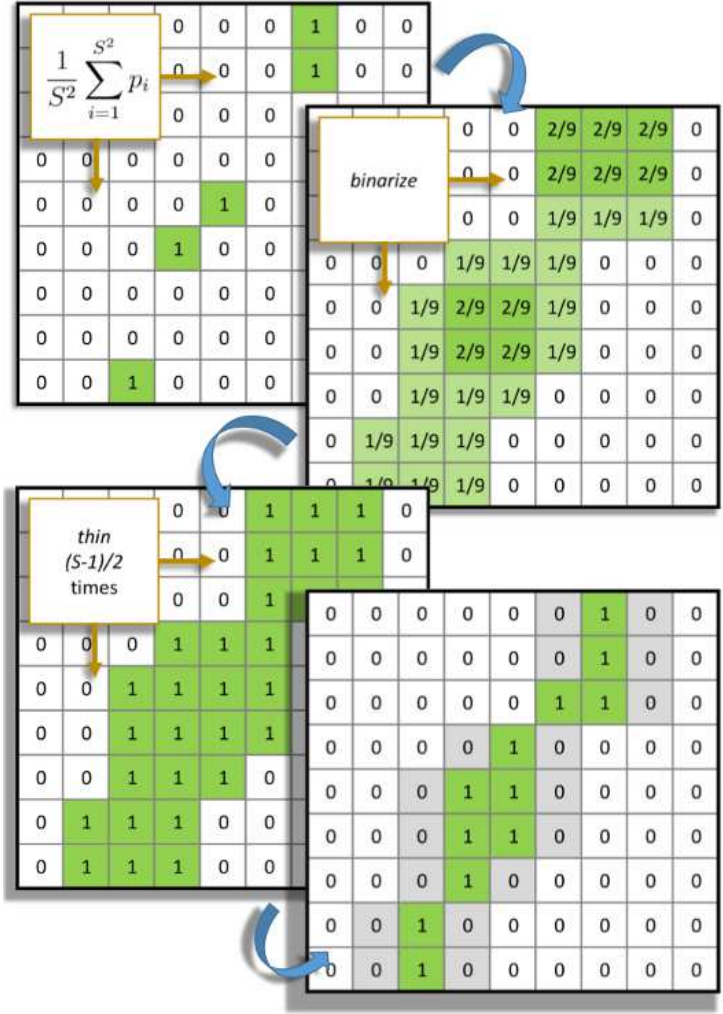}
\caption{Edge reconstruction. The $ED$ filter computes the edge density in the neighborhood $S\mbox{x}S$ of each pixel $p_i$; the effect is to blur the edge pixels through the filter area, forming a cluster where the gaps are filled. Next, binarization is applied to separate the edge cluster from the empty surroundings. A thinning operation is performed $(S-1)/2$ times in order to readjust the cluster shape; the most outer pixels are removed without causing new gaps.} 
\label{fig_filter_thin}
\end{figure}

Although many pixels in the resulting image represent edge discontinuities, the local ED in these pixels is not null. The effect of the filtering process is to blur the edges through the filter area (Fig. \ref{fig_isec_outline}.e), causing many gaps to be filled. Image areas massively occupied by edges become clusters that contrasts with the empty surroundings. After binarizing the image (Fig. \ref{fig_isec_outline}.f), clusters are precisely distinguished from the empty regions. ISEC follows the idea that the boundaries that separate these two regions are convenient to compose superpixels contours. 

The ED filter also causes a dilation on the clusters. That edge dilation has the side effect of driving the edges to lose their intrinsic adherence to the real object contours. This issue is coped by applying a thinning operation on the edge clusters in order to readjust their shape (Fig. \ref{fig_isec_outline}.g). The thinning is a morphological operation that removes the most external pixels in a segment of a binary image towards a line with the thickness of a single pixel -- not allowing for new discontinuities. Since the ED filter expands the clusters by $S-1$ pixels, the thinning operation needs to be performed $(S-1)/2$ times in order to readjust the cluster borders to correctly match the real object contours (considering that this operation performs symmetrically in both sides of the segments). Figure \ref{fig_filter_thin} illustrates, at the pixel level, how this filtering-binarizing-thinning process accomplishes the edge reconstruction.

After reshaping the edge clusters, the next step is to extract the cluster borders (Fig. \ref{fig_isec_outline}.h). In order to save processing time, this step is done by using look-up tables (LUT). These LUT's allow for fast pixel-level operations in a binary image, by replacing more expensive run-time calculations with simple searches for precomputed values in memory. To extract the cluster borders, we search for 1's pixels that have at least one 0 in their 3$\times$3 neighborhood. Before the searching, the look-up table has been fetched with all the 512 possible patterns that can be found in the 3$\times$3 neighborhood of the pixel in the binary image. At the end of each iteration, the set of extracted borders is stored. Additionally, the edge detection thresholds are incremented and the order, $S$, of the ED filter is reduced. The goal is to make the filter to better fit into the smaller parts of the objects at the next iteration.

\subsection{Superpixel refinement} \label{subsec_postproc}

When the segments provided by each previous iteration are assembled together, some borders eventually end up being adjacent, resulting in thick lines (Fig. \ref{fig_isec_outline}.i). Also, some segments can be too small or even individual pixels. The purpose of this superpixel refinement stage is to use morphological operations to remove such irregularities, improving the final result (Fig. \ref{fig_isec_outline}.j). Figure \ref{fig_isec_outline}.k shows the generated superpixels overlapped with the input image; some parts are zoomed to highlight segmentation details. The proposed superpixel segmentation method is summarized in Algorithm \ref{alg:spx}.

\begin{algorithm}[ht] 
        \caption{Iterative over-Segmentation via Edge Clustering}
        \label{alg:spx}
        \begin{algorithmic}[1]
        \small
        \Require Image $I$; size $S$ of the filter; lower and higher thresholds $[t,T]$
        \Ensure Labeled image containing superpixels $SPX$
        \For{each pixel $p_{i}$ in $I$} \do \\
         		\State $G$ $\leftarrow$ $argmax(gradient(p_{i\left \{ R \right \}},p_{i\left \{ G \right \}},p_{i\left \{ B \right \}}))$
		\EndFor
        \State $SPX$ $\leftarrow$ $0$
         \For{k = t to T} \do \\
         	\State $E$ $\leftarrow$ \textit{edgeSelection($G$,$k$)}
         	\For{each pixel $p_{i}$ in a region $S$$\times$$S$ of $E$} \do \\
         		\State $ED$ $\leftarrow$ $\frac{1}{S^2}\sum_{i=1}^{S^2}p_{i}$ \Comment{Edge Density}
         	\EndFor
         	\State \textit{Binarize $ED$}
         	\State \textit{Perform thinning of $ED$, $((S-1)/2)$ times}
         	\State $C$ $\leftarrow$ \textit{clusterContourExtraction($ED$)}
         	\State $SPX$ $\leftarrow$ \textit{OR($SPX$,$C$)}         				
			\If{$S > $\textit{minimum filter order}}
				\State $S$ $\leftarrow$ $S-2$
			\EndIf
	     \EndFor
	     \State \textit{Merge adjacent contours and remove isolated pixels in $SPX$}
         \State \textit{Label $SPX$}
        \end{algorithmic}
\end{algorithm}

\section{Experimental analysis} \label{sec_experiments}

\subsection{Methodology}

Most of the recent methods for superpixel generation rely on user-defined number of superpixels. That fact drove to performance metrics, which are computed as function of the number of superpixels, such as: \textbf{video-based metrics} -- motion undersegmentation error (MUSE) and motion discontinuity error (MDE) (see Section \ref{subsec_optFlowTest}), and \textbf{image-based metrics} -- undersegmentation error (UE) and boundary recall (BR) (see Section \ref{subsec_bsdTest}). For a given dataset, the usual evaluation methodology consists in running the methods on all the images repeatedly, each time generating a segmentation set with a different number of superpixels. The evaluation metrics are computed for all the images in each segmentation set, and averaged by the total of images. The results are usually plotted in a graph, putting the number of superpixels against the value of the respective metric. This methodology imposes constraints to evaluate methods such as ISEC, whose the number of superpixels depends on the image. In order to make ISEC be comparable to the other methods, the evaluations were run by varying the step in which the ISEC's edge detection thresholds are incremented. By setting a small step, the thresholds are increased more slowly, resulting in more edges being selected and, consequently, more superpixels. Conversely, the bigger the steps, the fewer the superpixels. It is noteworthy that such procedure only affects the number of superpixels indirectly. This procedure achieves an average number of superpixels on a given data set, although this number can considerably vary for each individual image.

Experimental evaluation also includes assessing performance of the evaluated methods based on an existing \textbf{application on video object segmentation} (see Section \ref{subsec_videoSegTest}). The methods are evaluated according to the overall performance achieved by the application to segment a video data set. For that video segmentation task, superpixel generation works as an independent module, allowing for the evaluation of different superpixel methods compared in the experiment.

\subsection{Video-based evaluation metrics} \label{subsec_optFlowTest}

In \cite{SPXBenchmark}, a benchmark on video segmentation is proposed to evaluate superpixel methods. Primarily, the benchmark provides completely data-driven metrics, based on optical flow. The motivation is to avoid the ground truth biasing due to eventual semantic knowledge inserted by human-made annotations. To better understand, suppose evaluating an object segmentation provided by a given method. Although separating the object from the background and surroundings is an obvious requirement, it is not always clear which subparts of the object itself need to be segmented apart. As humans intrinsically rely on contextual information when doing that task, metrics based on manually annotated data may be imprecise. Using optical flow as ground truth guarantees that only information contained in the image itself is used in the evaluation process.

The benchmark provides two metrics, based on the following criteria:

\begin{itemize}
\item \textit{Stability} - Evaluates the capability of the segmentation method to find the same regions or object boundaries regardless of image changes. The metric to assess this criterion is called MUSE, which measures the stability of a segmentation between two frames, $I_{1}$ and $I_{2}$, and is given as

\begin{equation}
	MUSE=\frac{1}{N}\left [ \sum_{a\in L_{1}^{F}} \left ( \sum_{b\in L_{2}:a\cap b\neq 0} min(b_{in},b_{out}) \right ) \right ] \, ,
\label{eq:muse}
\end{equation}
where $N$ is the total number of pixels, $L_{1}^{F}$ is a segmentation of $I_{1}$ in which the labels were transformed by using optical flow into the pixels of $I_{2}$; $L_{2}$ is the segmentation of $I_{2}$. Individual superpixels belonging to these segmentations are, respectively, $a$ and $b$; $b_{out}$ is the number of pixels of $b$ that are outside $a$, while $b_{in}$ is the number of pixels that are in $a\cap b$. The rationale is to segment subsequent frames, and then use the ground-truth optical flow to transform the first segmentation into the view of the second image, making the segmentations comparable. MUSE measures how well one segmentation can be reconstructed from the other.

\item \textit{Discontinuity} - Evaluates how well the generated superpixels can represent motion discontinuities in the image sequence. For instance, motion differences between moving foreground objects and the background, or among parts of the same object. The metric, called MDE, is defined as

\begin{equation}
	MDE=\frac{1}{\sum_{i}\sum_{j}\left \| \nabla F(i,j) \right \|_{2}}\sum_{i}\sum_{j}\left \| \nabla F(i,j) \right \|_{2}.D(B(i,j))\, ,
\label{eq:mde}
\end{equation}
where $F$ is the ground-truth optical flow between two frames, $B$ is the superpixel segmentation of one of these frames, and $D(B)$ is the distance transform of $B$ which contains, for each pixel, the distance to the nearest segment boundary. MDE accumulates a penalty over all image pixels, which is the product of the magnitude of motion discontinuity at this pixel and its distance to the next segment border (normalized by the total amount of motion in the image). Please, refer to \cite{SPXBenchmark} for more details on the metric.

\end{itemize}

\begin{figure}[!t]
\centering
\includegraphics[width=\textwidth]{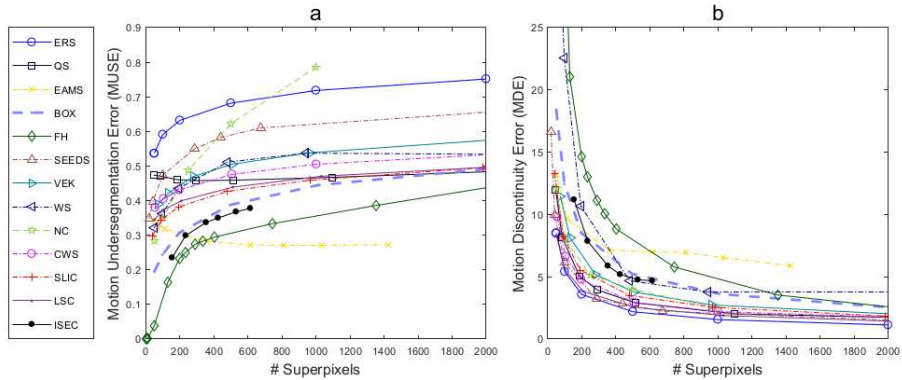}
\caption{Video-based evaluation results on Sintel data set. (a) MUSE measures the segmentation stability among the frames; methods that over-segment homogeneous regions in the image perform worse in this metric. (b) MDE evaluates how well the motion discontinuities present in the video sequence are represented by the generated superpixels; methods that provide segmentations where the superpixels are distributed by the whole image area can better represent motion discontinuities. ISEC is the only method to perform better than the baseline (BOX) in both MUSE and MDE, indicating that the proposed method has suitable features for video segmentation.}
\label{fig_sintel_plot}
\end{figure}

The benchmark uses the Sintel data set \cite{SintelDataset}, which has 23 computer rendered scenes, each one with 20 to 50 color images of size $1024\times436$. The ground-truth optical flow has been directly extracted from the data used for rendering. The usage of computer rendered images allows for the presence of denser optical flow data in Sintel compared to data sets based on natural images. This denser optical flow data supplies ground truth information for basically all image pixels, even when some pixels are occluded among a sequence of frames.   

We compare ISEC performance on Sintel data set with the methods: ERS \cite{ERS}, QS \cite{QS}, EAMS \cite{EAMS}, FH \cite{FH}, SEEDS \cite{SEEDS}, VEK \cite{VEK}, WS \cite{WS}, NC \cite{NC}, CWS \cite{CWS}, SLIC \cite{SLIC} and LSC \cite{LSC}.  Also, a simple regular grid segmentation (BOX) has been included for baseline comparison. For all the compared methods, the implementations are based on publicly available code. The results are illustrated in Fig. \ref{fig_sintel_plot}.

MUSE penalizes segments for which counterparts are not found in subsequent frames. Therefore an increase of the error with respect to the number of superpixels is expected, since by raising this number, homogeneous image regions are further segmented. Such homogeneous regions do not offer gradient support for the segmentation process, entailing a lot of variations in the segments among the frames, which contributes for instability. That is the reason why methods like EAMS, FH and ISEC, which do not rely on a grid-like segmentation, perform best on this stability criterion. In turn, the other methods present results that are next or even worse than the baseline BOX, as showed in Fig. \ref{fig_sintel_plot}.a.

For MDE, an opposite behavior is expected, that is, the greater the number of superpixels, the lower the error. This is because by increasing the image partitioning, more boundaries are produced. Hence it is more likely that some of these boundaries lie on motion discontinuities regions in the image. According to \cite{SPXBenchmark}, MDE is complementary to MUSE. This becomes evident in the plots of Fig. \ref{fig_sintel_plot}.b, where the methods appear in an inverted order compared to the previous MUSE evaluation. In MDE evaluation, methods such as SLIC, ERS, SEEDS, LSC, VEK, NC, CWS and QS, which provide grid-like segmentations (and consequently more distributed boundaries along the image area), present the best results. EAMS and FH, on the other hand, perform poorly on this metric.

\begin{figure}[!t]
\centering
\includegraphics[width=\textwidth]{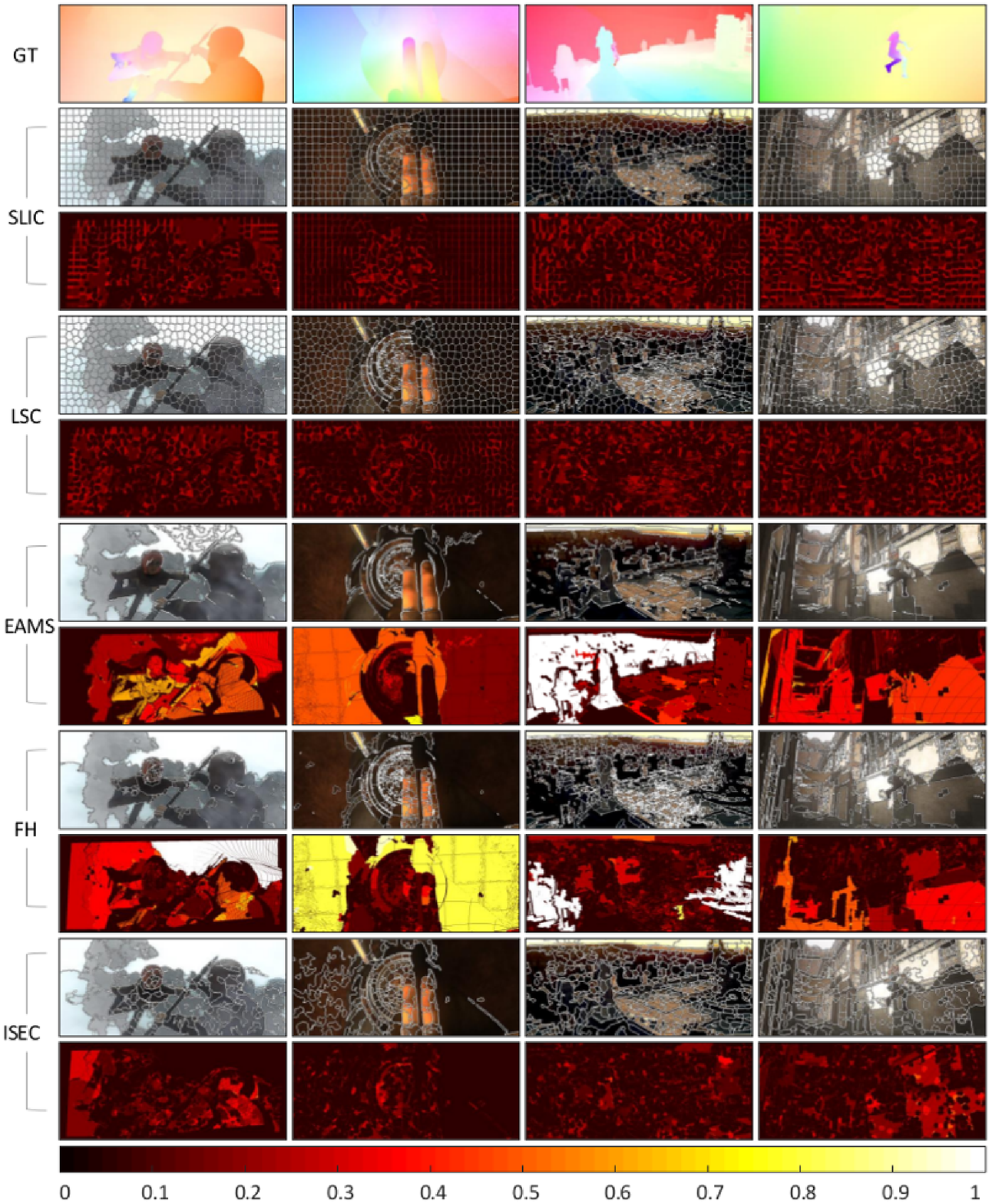}
\caption[LoF]{Visual comparison of superpixels generated on Sintel images. The first row depicts the ground-truth optical flow, according to the gradient map \includegraphics[height=\baselineskip]{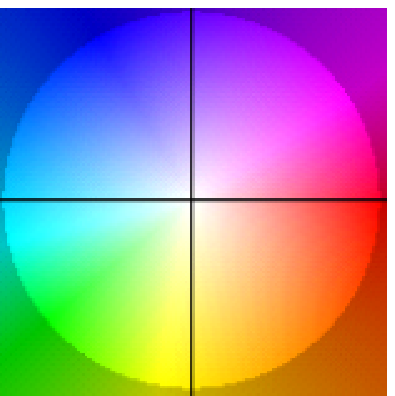}, where the color and intensity represent, respectively, the orientation and magnitude. For each method named on the left board, the rows show the generated superpixels (top row) and the MUSE (bottom row), which is represented in heat map, according to the scale showed at the bottom of the Figure.}
\label{fig_sintel_images}
\end{figure}

It is remarkable that on the Sintel data set, ISEC is the only method to perform better than the BOX in both MUSE and MDE. This result tackles the problem pointed out by \cite{SPXBenchmark}: methods that perform well on one criterion often show problems with the other, that is, stability or motion discontinuity representation. In words, the results indicate that our method provides a balance between segmentation stability and accuracy for video segmentation, filling the gap emphasized by the benchmark. Figure \ref{fig_sintel_images} shows a visual comparison of Sintel segmentations generated by ISEC and four other methods: SLIC and LSC, in which the number of superpixels is user-defined; and EAMS and FH, in which the number of superpixels is adaptive.

\subsection{Image-based metrics} \label{subsec_bsdTest}

\begin{figure}[!t]
\centering
\includegraphics[width=\textwidth]{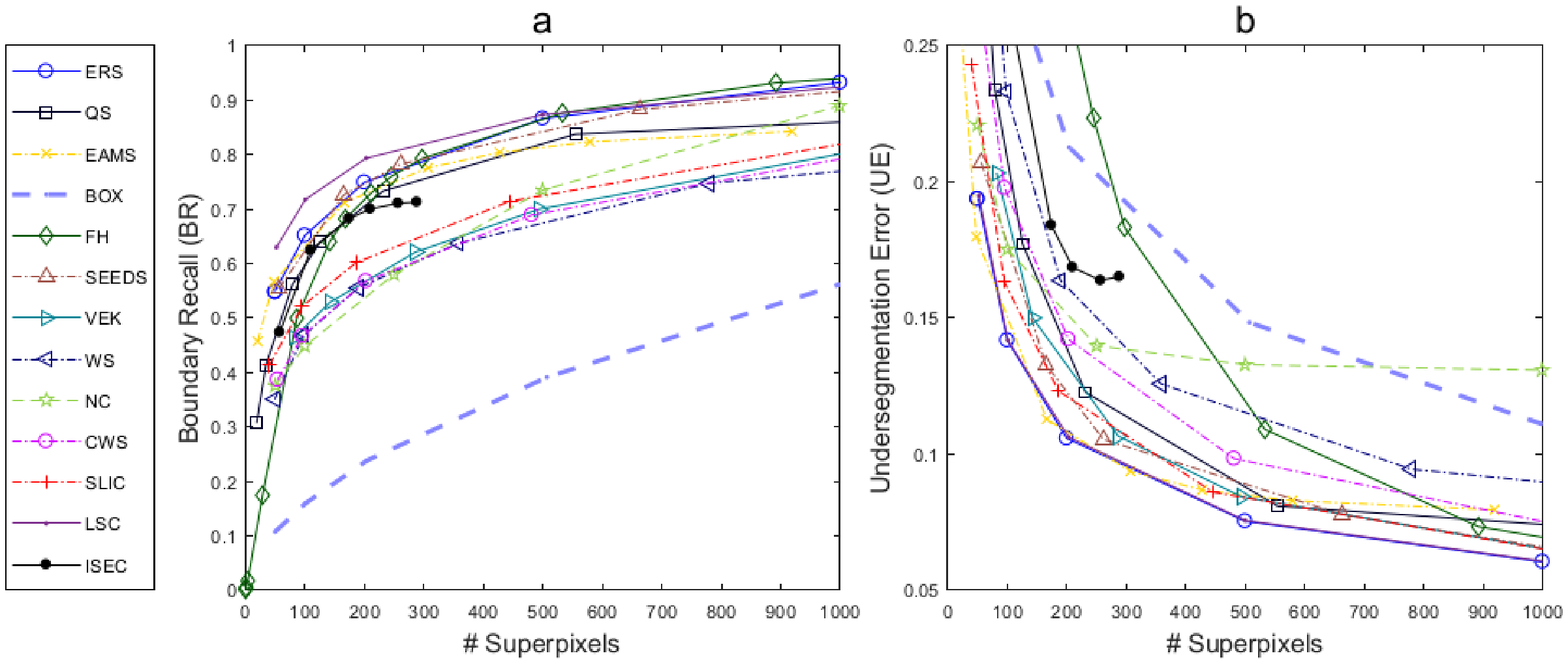}
\caption{Image-based evaluation results on BSD data set. (a) BR measures the matching between superpixel boundaries and the ground-truth, which is comprised of human-annotated object contours. The plot shows that ISEC is in line with the group formed by the methods with better performance regarding BR, only being considerably surpassed after the other methods get an excessive number of superpixels. (b) UE penalizes superpixels that overlaps more than one object in the image; methods that try to avoid the over-segmentation of homogeneous regions in the image, such as ISEC and FH, are prone to be further penalized by this metric.
}
\label{fig_bsd_ue_br_plot}
\end{figure}

Berkeley Segmentation Database (BSD) \cite{Berkeley} is by far the most used to evaluate superpixel methods over image-based metrics. This data set provides 500 color images split into 200 for training, 100 for validation and 200 for test, as well as the respective human annotated ground-truth. Usually segmentation performance on BSD is assessed from the widely known metrics BR and UE. Although our method is geared towards video segmentation, for the sake of completeness we have included experiments on BSD. The evaluations were run on the 200 test images, comparing the same algorithms as the previous Sintel test. The results are presented in Fig. \ref{fig_bsd_ue_br_plot}.

\begin{figure}[!t]
\centering
\includegraphics[width=\textwidth]{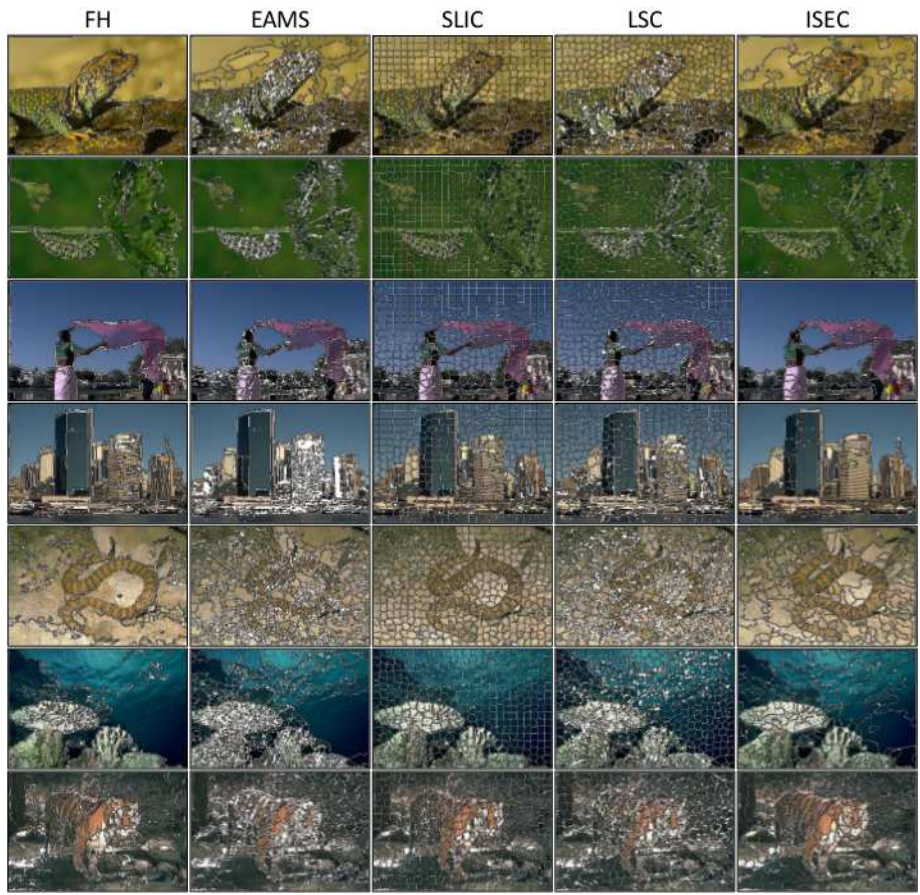}
\caption{Visual comparison of superpixels generated on BSD images. For SLIC and LSC, whose the number $k$ of superpixel is user-defined, the presented examples were generated with $k=500$. For FH, EAMS and ISEC, whose $k$ is adaptive, the examples were generated by adjusting the internal parameters of these methods in order to provide $k\approx 500$ (in average, considering the entire BSD data set).}
\label{fig_bsd_images}
\end{figure}

Figure \ref{fig_bsd_ue_br_plot}.a illustrates results using BR, which reveals that all the methods far outperform the BOX in respect to adherence to real image boundaries. Besides, the evaluated methods appear to form two main groups according to their performance: the best performance group, where the methods reach about 70\% of BR using a maximum of 200 superpixels; and the second best performance group, where the algorithms need more than twice that number of superpixels to reach the same performance. ISEC is in line with the former group, only being considerably surpassed after the other methods get an excessive number of superpixels. A visual comparison of superpixel segmentations on BSD data set is showed in Fig. \ref{fig_bsd_images}. Besides ISEC, the the visual comparison includes segmentations provided by SLIC and LSC, whose the number of superpixels is user-defined, and FH and EAMS, whose that number is automatically controlled.

Regarding the UE (see Fig. \ref{fig_bsd_ue_br_plot}.b), although ISEC performance is always better than the BOX, it is not as good as most of the other methods, except for FH. That is the drawback of trying to produce a segmentation that is fully supported by the image content. Despite the proposed method presents strategies to deal with eventual lacks of information (see ISEC edge reconstruction in Section \ref{sec_propMethod}), eventually an object with a too long and smooth contour can lead to issues. That is why recent superpixel methods rely on grid-like segmentation strategies. However, in the case of video segmentation, sometimes it is worth dealing with a higher UE in favor of avoiding to segment image regions unnecessarily, as well as to obtain more stable superpixels.

\begin{figure}[!t]
\centering
\includegraphics[width=0.6\textwidth]{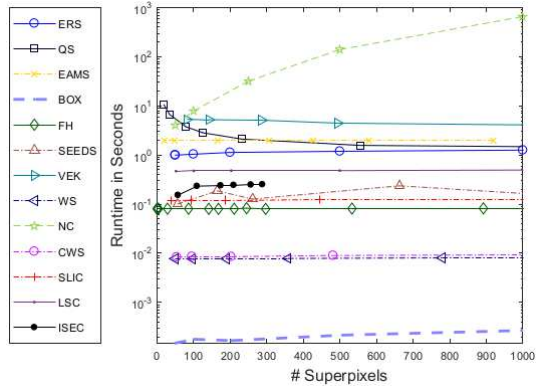}
\caption{Computational time. WS and CWS are the fastest methods, taking less than 10 ms per image; FH, SLIC, SEEDS, ISEC and LSC come next, spending around a few hundred milliseconds; ERS, EAMS, QS, VEK spend more than a second; and NC is not linear, taking several minutes to generate the superpixels even for resized images. Among the fastest methods, ISEC is the only one implemented in Matlab, while the other methods are implemented in C/C++.}
\label{fig_bsd_rt_plot}
\end{figure}

Finally, results of \textbf{computational time} on BSD are illustrated in Fig. \ref{fig_bsd_rt_plot}. The plots show that our method is one of the fastest. By order: WS and CWS come first, taking around 10 ms per image; followed by FH and SLIC ($\approx$ 100 ms), SEEDS and ISEC ($\approx$200 ms), and LSC ($\approx$500 ms). The remaining methods spend over a second per image, drawing attention the scalable time consumption of NC (hundreds of seconds), even for resized images. Also note that, while all the fastest methods were implemented in C/C++, ISEC was coded in Matlab, taking room for speed-ups.

\subsection{Application on video object segmentation} \label{subsec_videoSegTest}

This evaluation is based on a framework for video object segmentation proposed in \cite{FastVideo}. This framework is divided into two main steps: (i) optical flow is used to estimate an initial foreground segmentation based on motion, and (ii) appearance cues are extracted from the segments, relying on superpixels to obtain refined object contours. Since the superpixel generation is an independent module in this video object segmentation framework, it is possible to evaluate the impact of using different methods for superpixel generation.

Tests were run on the SegTrack data set \cite{SegTrackDataset}. This data set provides 6 video sequences (monkeydog, girl, birdfall, parachute, cheetah and penguin), along with the object annotation for each frame. Following \cite{FastVideo}, the penguin video has been discarded due to inconsistencies in the ground-truth, remaining a total of 202 frames. Table \ref{segVidTable} presents the results of using ISEC, as well as other five representative methods selected from experiments on Sintel and BSD. Table columns summarizes: (1) the evaluated methods; (2-6) the average number of mislabeled pixels for each SegTrack video; (7) the average error, given by the total of mislabeling divided by the number of frames; (8) the average number of superpixels generated for each frame; and (9) the total time spend to segment the entire dataset.

\begin{table}[t]
\centering
\resizebox{\textwidth}{!}{%
\begin{tabular}{@{}lcccccccc@{}}
\toprule
\multirow{2}{*}{Method} & \multicolumn{5}{c}{SegTrack Videos} & \multirow{2}{*}{\begin{tabular}[c]{@{}c@{}}Average\\ error\end{tabular}} & \multirow{2}{*}{\begin{tabular}[c]{@{}c@{}}Average\\ \# Superp.\end{tabular}} & \multirow{2}{*}{\begin{tabular}[c]{@{}c@{}}Total\\ Time (s)\end{tabular}} \\ \cmidrule(lr){2-6}
 & Birdfall & Cheetah & Girl & Monkey & Parachute &  &  &  \\ \midrule
FH & - & 1462 & - & - & 764 & 1113 & 49 & 12.6 \\
EAMS & - & 915 & 2536 & 328 & 19062 & 5710 & 324 & 274.9 \\
SLIC & 345 & 902 & 2954 & 402 & 452 & 1011 & 1500 & 20.7 \\
ERS & 288 & 873 & 1802 & 1439 & 305 & 941 & 700 & 196.0 \\
LSC & 217 & 913 & 1861 & 350 & 273 & 723 & 500 & 62.2 \\
ISEC & 268 & 973 & 2324 & 390 & 407 & 872 & 324 & 32.9 \\ \bottomrule
\end{tabular}%
}
\caption{Results on the video object segmentation experiment. The columns show: (1) the evaluated methods; (2-6) the average number of mislabeled pixels for each SegTrack video; (7) the average error, given by the total of mislabeling divided by the number of frames; (8) the average number of superpixels generated for each frame; and (9) the total time spend to segment the entire dataset. The missing data (columns 2, 4, 5 for FH, and column 2 for EAMS) are cases where the video object segmentation failed to converge due to degenerate superpixels. The results indicate LSC and ISEC as the most efficient methods: low error rate with a small number of superpixels. Regarding the processing time, ISEC is twice as fast as LSC.
}
\label{segVidTable}
\end{table}

\begin{figure}[!t]
\centering
\includegraphics[width=\textwidth]{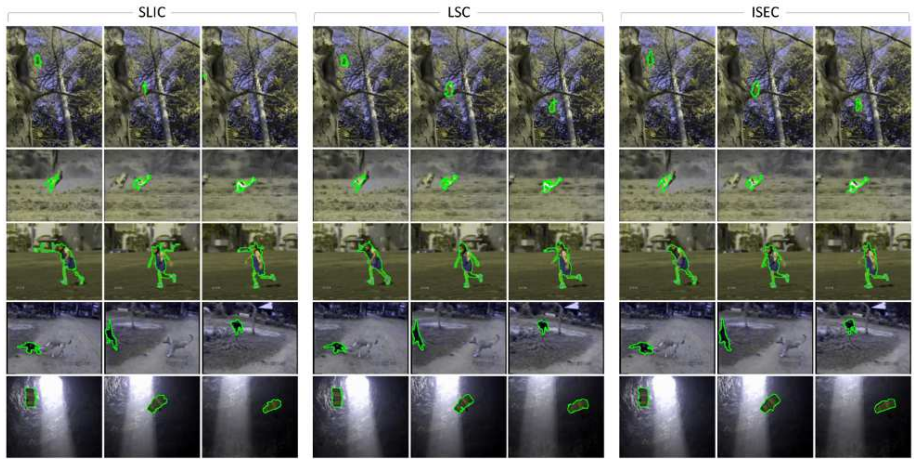}
\caption{Visual comparison of object segmentation on SegTrack sequences. From the top to the bottom, the rows show, for each method, three distinct shots of the video sequences named birdfall, cheetah, girl, monkey and parachute. In green, is the object segmentation achieved by \cite{FastVideo} when using the different superpixel methods, SLIC, LSC and ISEC.}
\label{fig_segtrack_images}
\end{figure}

The missing data (columns 2, 4, 5 for FH, and column 2 for EAMS) are cases where the video object segmentation failed to converge due to degenerate superpixels. In the case of methods based on a user-defined number of superpixels (SLIC, ERS and LSC), the tests were started with a small number of superpixels, being this number increased until achieving convergence. Results show that despite FH is very fast and generates a small number of segments, this method is not able to support video segmentation in most of the cases. In the cases where FH was successful, this method presented the second worst average error. EAMS, in turn, besides the highest error rate, is by far the slowest method, spending 274 seconds to segment all the videos. Intermediate results were reached by SLIC and ERS. While SLIC needs more than twice the number of superpixels than the other methods to converge, ERS is also too slow (the second in computational time). The best results were achieved by LSC and ISEC. The former presents low error rate (only 723 pixels in average) with few generated superpixels (500 per frame), being moderate in terms of speed. ISEC gets low error (872 pixels in average, the second best), generating fewer superpixels per image (324) in average, being twice as fast as LSC (around 33 against 62 seconds). Figure \ref{fig_segtrack_images} compares some examples of object segmentation provided by the video segmentation framework when using SLIC, LSC and ISEC.

\section{Conclusions} \label{sec_conclusions}

We introduced this paper with a discussion about whether is convenient or not, for a superpixel method, to rely on a fixed number of superpixels. Our hypothesis was pointed in the direction that automatically controlling this number could be suitable for some tasks, specially regarding video segmentation. To test that hypothesis, we compared the proposed method with several state-of-the-art methods through a set of experiments designed to evaluate different aspects of superpixel segmentation. The evaluation using video-based metrics reinforced the idea that ISEC approach is suitable for video segmentation; the proposed method was the best in providing accurate representation of motion discontinuities, while preserving segmentation stability in video sequences. Indeed, these results were corroborated by another experiment, this time involving an application for video object segmentation. By this latter experiment, ISEC proved to be one of the most efficient methods, leading the application to accomplish the second best segmentation performance, with the smallest number of superpixels and spending half of the time of the superpixel method that presented the best performance in this experiment. On the other hand, the experiment using image-based metrics indicated that approaches based on a fixed number of superpixels are the better choice for static image segmentation, since these types of methods are able to provide high boundary adherence, while preventing undersegmentation failures.

\bibliography{mybibfile}

\end{document}